\documentclass{article}

\PassOptionsToPackage{numbers, compress, sort}{natbib}

\usepackage[preprint]{neurips_2025}

\usepackage[utf8]{inputenc} 
\usepackage[T1]{fontenc}    
\usepackage{hyperref}       
\usepackage{url}            
\usepackage{booktabs}       
\usepackage{amsfonts}       
\usepackage{nicefrac}       
\usepackage{microtype}      
\usepackage{xcolor}         
\usepackage{amsmath}
\usepackage{graphicx}
\usepackage{multirow}
\usepackage{float}

\title{MMbeddings: Parameter-Efficient, Low-Overfitting Probabilistic Embeddings Inspired by Nonlinear Mixed Models}

\author{%
  Giora Simchoni \\
  Department of Statistics\\
  Tel Aviv University\\
  Tel Aviv, Israel, 69978 \\
  \texttt{gsimchoni@tauex.tau.ac.il} \\
   \And
   Saharon Rosset \\
   Department of Statistics\\
   Tel Aviv University\\
   Tel Aviv, Israel, 69978 \\
   \texttt{saharon@tauex.tau.ac.il} \\
}

\begin{document}

\maketitle

\begin{abstract}
  We present MMbeddings, a probabilistic embedding approach that reinterprets categorical embeddings through the lens of nonlinear mixed models, effectively bridging classical statistical theory with modern deep learning. By treating embeddings as latent random effects within a variational autoencoder framework, our method substantially decreases the number of parameters—from the conventional embedding approach of cardinality $\times$ embedding dimension, which quickly becomes infeasible with large cardinalities, to a significantly smaller, cardinality-independent number determined primarily by the encoder architecture. This reduction dramatically mitigates overfitting and computational burden in high-cardinality settings. Extensive experiments on simulated and real datasets, encompassing collaborative filtering and tabular regression tasks using varied architectures, demonstrate that MMbeddings consistently outperforms traditional embeddings, underscoring its potential across diverse machine learning applications.
\end{abstract}

\section{Introduction}
\label{sec:intro}
Categorical features play a critical role in a variety of predictive modeling tasks, yet efficiently managing these features becomes increasingly challenging as their cardinality grows \citep{lmmnn_neurips}. For instance, collaborative filtering methods commonly deal with categorical features such as user and item identifiers, each with cardinality in the thousands or even millions, resulting in extremely high-dimensional representations and posing substantial computational and statistical challenges \citep[see, e.g.,][]{NeuralCF}. One-hot encoding (OHE) represents each category of a categorical feature of \(q\) levels as a separate binary indicator feature, leading to \(q\) additional features. When cardinality is high, this approach results in a computationally burdensome and inefficient representation that fails to encode similarity between related categories and significantly increases the risk of overfitting \citep{Hancock2020}.

Entity embeddings represent categorical features by mapping each of the \(q\) distinct categories into a continuous \(d\)-dimensional vector space using a learned embedding matrix \(\mathbf{B} \in \mathbb{R}^{q \times d}\) \citep{guo2016entity, YoutubeEmbeddings}. The embeddings approach has become a ubiquitous modeling strategy, significantly improving performance over traditional encodings, with applications ranging from tabular and relational data in predictive modeling tasks \citep{Lin2019} to representing tokens in natural language processing \citep{mikolov2013} and powering recent state-of-the-art architectures such as transformers \citep{huang2020tabtransformer}. While this representation captures semantic relationships between categories and substantially reduces dimensionality compared to OHE, it could still introduce a considerable number of parameters, leading to high computational complexity and vulnerability to overfitting \citep{UnifiedEmbeddings_Neurips23}.

Although embedding layers seem like a modern machine learning innovation, categorical data embeddings have quietly existed within statistical models for decades, rooted implicitly through the use of random effects in linear and non-linear mixed models \citep{McCulloch2008}. Linear mixed models (LMM) distinguish between fixed effects (FE) -- standard model parameters estimated directly from data -- and random effects (RE), which are treated as random variables whose distributions are estimated instead. This distinction effectively handles high-cardinality categorical features or clustering structures within data \citep{lmmnn_jmlr}. For example, in medical research, treatment type (a carefully designed experimental condition) is modeled as a FE, while observational categorical features such as a patient's city -- with potentially thousands of unique levels -- are modeled as RE, typically assumed to follow a normal distribution with mean zero and unknown variance. Extending this idea, non-linear mixed models (NLMM) with \(d\)-dimensional RE assigned to each categorical unit can be viewed as representing these categories through latent embeddings drawn from probability distributions, thus offering a robust and theoretically grounded probabilistic interpretation of embeddings. This connection will be elaborated in Section~\ref{sec:background}.

Building upon this historical connection between embeddings and NLMM, we introduce MMbeddings -- named to explicitly highlight their foundation in ``mixed models embeddings.'' Our proposed approach leverages a novel variational autoencoder (VAE) architecture, explicitly modeling embeddings as RE within a VAE framework while minimizing a negative log-likelihood (NLL) loss. By treating embeddings as latent variables drawn from probability distributions, this architecture naturally imposes regularization, substantially reduces the number of parameters, and mitigates overfitting. Extensive experiments on both simulated datasets and real-world data demonstrate that MMbeddings achieve superior predictive performance, consistently outperforming traditional entity embeddings and standard encoding techniques across metrics such as MSE in regression and AUC in classification.

The remainder of this paper is structured as follows. Section~\ref{sec:background} reviews background concepts and discusses related methods in entity embeddings and mixed models. In Section~\ref{sec:mmbeddings}, we formally introduce our proposed method, MMbeddings, detailing its variational architecture and probabilistic mechanism. Extensive experiments demonstrating the effectiveness of MMbeddings compared to alternative methods, are presented in Section~\ref{sec:experiments}. Finally, Section~\ref{sec:conclusion} summarizes our contributions and suggests potential directions for future research.

\section{Background and related methods}
\label{sec:background}
NLMM have a rich history in statistical modeling, with the landmark paper by \citet{LindstromBates1990} introducing NLMM specifically for analyzing repeated measures data. One illustrative example from their seminal work involved repeated circumference measurements \((x_{ij}, y_{ij})\) of orange trees, where \(x_{ij}\) denotes the time of the \(i\)-th measurement for tree \(j\) (\(j = 1, \dots, q\) trees, \(i = 1, \dots, n_j\) measurements per tree). Lindstrom and Bates proposed the logistic NLMM:
\begin{equation}
\label{eq:Lindstrom}
    y_{ij} = \frac{\beta_1 + b_{1j}}{1 + \exp\left(-\frac{x_{ij} - (\beta_2 + b_{2j})}{\beta_3 + b_{3j}}\right)} + \varepsilon_{ij},
\end{equation}
where \(\beta_1, \beta_2, \beta_3\) represent FE and \(\varepsilon_{ij}\) random Gaussian noise. Specifically, \(\beta_1\) describes the average asymptotic circumference of the tree, \(\beta_2\) indicates the average time a tree reaches half its maximum circumference, and \(\beta_3\) acts as a scale parameter. Additionally, each tree \(j\) has associated RE \(b_{kj} \sim \mathcal{N}(0, \sigma^2_{b_k})\) for \(k = 1, 2, 3\), where \(\sigma^2_{b_k}\) are variance components estimated from data. Notably, these RE can be stacked into a matrix \(\mathbf{B}_{q \times 3}\), mapping each tree to a continuous latent embedding in a \(d=3\) dimensional space, closely aligning with the modern concept of embeddings. Lindstrom and Bates devised a two-step iterative maximum likelihood procedure: initially estimating FE and RE given initial variance components, then refining variance estimates using a Newton-Raphson approach to maximize the conditional log-likelihood of \(y | \mathbf{B}\). However, their method was developed in a relatively ``small data'' context (their original dataset contained only \(q = 5\) trees, each measured \(n_j = 7\) times), limiting scalability. Furthermore, their approach required explicitly specifying a parametric non-linear function. Our proposed method builds upon this foundational work, generalizing Lindstrom and Bates' model to handle large-scale data. Instead of explicitly defining a parametric relationship, we employ a deep neural network (DNN) within an encoder-decoder architecture to flexibly model complex, non-linear relationships. By minimizing a variational loss that serves as an upper bound to the NLL, our approach effectively unifies and modernizes Lindstrom and Bates' two-step iterative procedure, leveraging contemporary machine learning practices such as mini-batch processing and stochastic gradient descent (SGD).

Several recent approaches have attempted to integrate RE into DNN, with no specific focus on embeddings. MeNets \citep{Xiong_2019_CVPR} focus on random intercept models, specifically modeling a single categorical feature with a single RE, represented as \(y_{ij} = f(\mathbf{x}_{ij})+b_{0j} + \varepsilon_{ij}\), where \(b_{0j} \sim \mathcal{N}(0, \sigma^2_b)\). Other methods like sgGP \citep{Chen2020} and deep kernel learning \citep{DKL, SVDKL} integrate Gaussian processes into deep learning frameworks, particularly tailored to spatial datasets with measurements at \(q\) distinct locations. LMMNN \citep{lmmnn_jmlr} incorporates RE into DNN by decomposing inputs into separate FE features \(\mathbf{x}_{ij}\) and RE features \(\mathbf{z}_{ij}\), using a LMM-inspired formulation \(y_{ij} = f(\mathbf{x}_{ij})+g(\mathbf{z}_{ij})\mathbf{b}_{j} + \varepsilon_{ij}\), with Gaussian NLL loss. However, this strict separation limits identifiability when attempting to represent each categorical unit \(j\) with a \(d\)-dimensional RE vector \(\mathbf{b}_j\) (with \(d > 1\)), due to inherent identifiability issues highlighted in \citet{Lavielle2016}. A recent approach closely related to ours, which we term REbeddings \citep{Richman2024}, explicitly models embeddings as RE within an NLMM, minimizing a variational bound on the NLL. However, REbeddings parameterize embeddings using two separate embedding layers -- one each for the means and log-variances for all elements in \(\mathbf{B}_{q \times d}\) -- resulting in a prohibitive number of parameters (\(2Kqd\) parameters for \(K\) categorical features of cardinality \(q\) with embedding dimension \(d\)). In contrast, our proposed method leverages a neural encoder to output only \(d\) means and log-variances to produce posterior distributions for embeddings based on input features \(\mathbf{X}\) and outcomes \(\mathbf{y}\), requiring only \(M + 2d\) parameters (where \(M\) is the number of encoder parameters, typically much smaller than \(Kqd\)). Our method significantly reduces overfitting risks with improved predictive performance as demonstrated in Section~\ref{sec:experiments}.

Several alternative encoding techniques generate categorical embeddings without explicit training, such as feature hashing \citep{FeatureHashing}, mean-encoding \citep{kuhn2019feature}, and PCA-encoding \citep{liang2025}. Although computationally efficient, these heuristic methods typically yield inferior predictive performance compared to learned embeddings, as demonstrated in our experiments in Section~\ref{sec:experiments}.

A distinct branch of research has aimed to curtail the growing complexity and dimensionality of trainable embeddings. Hash Embeddings \citep{HashEmbeddings_NIPS2017} employ feature hashing within embedding layers to substantially decrease the parameter count, mitigating collisions through multiple hash functions and learned combinations. Similarly, Compositional Embeddings \citep{CompositionalEmbeddings} represent categories through combinations of shared sub-embeddings, significantly reducing the number of embedding parameters needed. Most notably, Unified Embeddings (UEmbeddings) \citep{UnifiedEmbeddings_Neurips23} generalize these ideas by using a unified embedding framework that blends hashing and compositional techniques, mapping categories from different features into a shared latent space through a single embedding matrix \(\mathbf{B}_{q_{U}\times d}\). While these methods effectively reduce parameter counts, they tend to be predominantly architecture-driven, relying on heuristic design choices without explicit underlying probabilistic or statistical models.

Recent years have witnessed considerable progress in advanced methods tailored to tabular data, particularly emphasizing categorical feature representations through embedding layers. Factorization machines \citep{FactorizationMachines} and their DNN variant, DeepFM \citep{DeepFM}, explicitly model feature interactions leveraging embeddings. Similarly, collaborative filtering methods \citep{CollaborativeFiltering} and their DNN extension, Neural Collaborative Filtering \citep{NeuralCF}, combine embeddings to capture complex user-item interactions. More recently, TabNet \citep{TabNet} introduced attention mechanisms to sequentially focus on relevant features, while TabTransformer \citep{huang2020tabtransformer} applies transformer architectures specifically to categorical data embeddings, effectively modeling interactions among categories. Principally, despite their methodological diversity, these sophisticated approaches uniformly rely on standard embedding layers. In Section~\ref{sec:experiments}, we directly compare some of these methods equipped with conventional embeddings against our proposed MMbeddings framework.

\section{Proposed method: MMbeddings}
\label{sec:mmbeddings}
Consider an observation \( y_{ij} \in \mathbb{R} \), representing the \( i \)-th data point within the \( j \)-th level of a single categorical feature that has \( q \) distinct levels (with \( j = 1, \dots, q \) and \( i = 1, \dots, n_j \)). Extending the formulation given in \eqref{eq:Lindstrom}, our model can be expressed more generally as:
\begin{equation}
\label{eq:MMbeddings_scalar}
    y_{ij} = f(\mathbf{x}_{ij}, \mathbf{b}_{j}) + \varepsilon_{ij},
\end{equation}
where \( \mathbf{x}_{ij} \in \mathbb{R}^p \) denotes the vector of additional features, \( \varepsilon_{ij} \sim \mathcal{N}(0, \sigma^2) \) is Gaussian noise, \(f\) is a non-linear function typically modeled with DNN, and \( \mathbf{b}_j \in \mathbb{R}^d \) is the RE vector, which we refer to as MMbeddings. Typically, \( \mathbf{b}_j \) is assumed to follow a multivariate normal distribution \( \mathcal{N}(\mathbf{0}, \mathbf{D}) \), with \( \mathbf{D} \) being a general covariance matrix. However, it often proves beneficial to assume \( \mathbf{D} \) is diagonal, implying each element \( b_{mj} \) of \( \mathbf{b}_j \) independently follows \( \mathcal{N}(0, \sigma_{m}^2) \), \(m = 1, \dots, d\).

To accommodate multiple categorical features, we expand our notation. For instance, with two categorical features, the observation can be written as \( f(\mathbf{x}_{ijl}, \mathbf{b}_{j}, \mathbf{b}_{l}) \), reflecting the inclusion of two MMbeddings vectors corresponding to each categorical feature. Generalizing further, for \( K \) categorical features, each with cardinality \( q_k \), we collect the corresponding MMbeddings vectors into \( q_k \times d_k \) embedding matrices \( \mathbf{B}_k \). Consequently, we succinctly represent our model in vectorized form:
\begin{equation}
\label{eq:MMbeddings_vectorized}
    \mathbf{y} = f(\mathbf{X}, \mathbf{B}_1, \dots, \mathbf{B}_K) + \pmb{\varepsilon},
\end{equation}
where \( \mathbf{y} \in \mathbb{R}^n \) is the response vector, \( \mathbf{X} \) is the \( n \times p \) matrix of covariates, and \( \pmb{\varepsilon} \in \mathbb{R}^n \) is a Gaussian noise vector.

Consider again the scenario with a single categorical feature described in \eqref{eq:MMbeddings_scalar}, and denote the vector of observations within cluster \( j \) as \( \mathbf{y}_j = [y_{1j}, \dots, y_{n_j,j}] \). Directly computing the marginal log-likelihood of \( \mathbf{y}_j \), obtained by integrating out the MMbeddings \( \mathbf{b}_j \) -- specifically, \( \log p(\mathbf{y}_j) = \log \int p(\mathbf{y}_j|\mathbf{b}_j)p(\mathbf{b}_j)d\mathbf{b}_j \) -- is computationally intractable. A standard solution from variational Bayes \citep[see, e.g.,][]{VAE2013} introduces an approximating distribution \( q(\mathbf{b}_j|\mathbf{y}_j) \) for the posterior of \( \mathbf{b}_j \). This approximation yields the evidence lower bound (ELBO), providing a tractable lower bound to the log-likelihood:
\begin{equation}
    \log p(\mathbf{y}_j) \ge \text{ELBO} := \mathbb{E}_{q}\left[\log p(\mathbf{y}_j|\mathbf{b}_j)\right] - D_{KL}\left[q(\mathbf{b}_j|\mathbf{y}_j) \| p(\mathbf{b}_j)\right].
\end{equation}
Maximizing the ELBO thus corresponds to simultaneously maximizing the expected conditional log-likelihood \( \mathbb{E}_q[\log p(\mathbf{y}_j|\mathbf{b}_j)] \) and minimizing the Kullback–Leibler divergence between the approximate posterior \( q(\mathbf{b}_j|\mathbf{y}_j) \) and the prior distribution \( p(\mathbf{b}_j) \). The conditional distribution \( p(\mathbf{y}_j|\mathbf{b}_j) \) is Gaussian  \( \mathcal{N}(f(\mathbf{x}_j, \mathbf{b}_j), \sigma^2 \mathbf{I}_{n_j}) \) as indicated by \eqref{eq:MMbeddings_scalar}. In practice, a single sample drawn from \( q(\mathbf{b}_j|\mathbf{y}_j) \) is typically sufficient to estimate the expectation. Moreover, following the mean-field variational inference framework, we assume \( q(\mathbf{b}_j|\mathbf{y}_j) \) factorizes over dimensions, i.e., \( q(\mathbf{b}_j|\mathbf{y}_j) = \prod_{m = 1}^d q_m(b_{mj}|\mathbf{y}_j) \). The distribution of each factor \( q_m(b_{mj}|\mathbf{y}_j) \) is assumed Gaussian with variational parameters \( \mu_{mj} \) and \( \tau^2_{mj} \), that is, \( q_m(b_{mj}|\mathbf{y}_j) = \mathcal{N}(\mu_{mj}, \tau_{mj}^2) \). These variational parameters are fitted during training.

\begin{figure}
    \centering
    \includegraphics[width=0.6\linewidth]{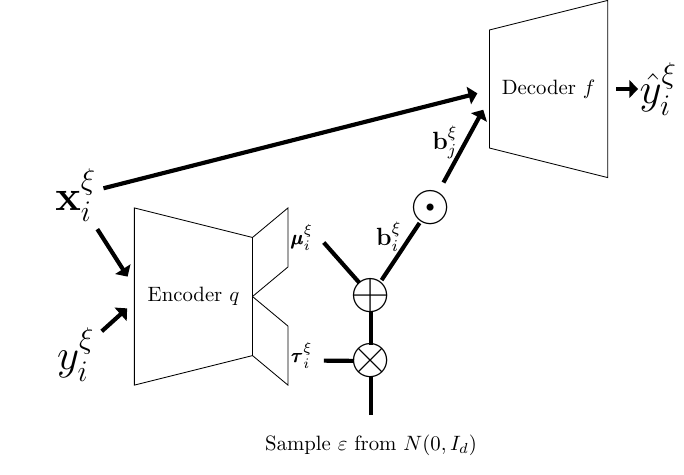}
    \caption{MMbeddings scheme for a single categorical feature, combining an encoder to parameterize the variational posterior \( q(\mathbf{b}_j|\mathbf{y}_j) \) and a decoder modeling the nonlinear function \( f(\mathbf{x}_{ij}, \mathbf{b}_j) \).}
    \label{fig:MMbeddings_scheme}
\end{figure}

Figure~\ref{fig:MMbeddings_scheme} illustrates the overall architecture of MMbeddings. As commonly adopted in the VAE framework, we utilize a neural network encoder to parameterize the approximate posterior distribution \( q(\mathbf{b}_j|\mathbf{y}_j) \). For each observation \((\mathbf{x}_{i}, y_{i})\) of cluster \(j\), this encoder produces vectors of \(d\) means \( \pmb{\mu}_{j} \) and \(d\) log-variances \( \log \pmb{\tau}_{j}^2 \) for the latent embedding \( \mathbf{b}_j \). Using the reparameterization trick, we sample \( \mathbf{b}_j \) from this variational posterior to maintain differentiability during training. A separate neural network decoder then models the nonlinear relationship \( f \), taking as input the sampled embedding \( \mathbf{b}_j \) along with the covariates \( \mathbf{x}_{i} \). The decoder architecture can vary widely -- including a simple multilayer perceptron (MLP), neural collaborative filtering, or transformer -- as we demonstrate in Section~\ref{sec:experiments}. The decoder, crucially, is later employed independently at inference for prediction tasks.

The negative ELBO serves as our training objective. Simplifying this, for an individual observation \((\mathbf{x}_{i}, y_{i})\) from cluster \(j\), the negative ELBO can be explicitly written as:

\begin{equation*}
    -ELBO_{i} = \frac{1}{2}\log(2\pi\sigma^2) + \frac{1}{2\sigma^2} (y_{i} - f(\mathbf{x}_{i}, \mathbf{b}_j))^2 +  \frac{1}{2} \sum_{m = 1}^d\left[-1 - \log\tau^2_{mj} + \log\sigma^2_{m} + \frac{\mu^2_{mj}}{\sigma^2_{m}} + \frac{\tau^2_{mj}}{\sigma^2_{m}}\right].
\end{equation*}

When moving to a minibatch \(\xi\) of size \( n_\xi \), comprising observations \( \left[(\mathbf{x}^{\xi}_1, y^{\xi}_1), \dots, (\mathbf{x}^{\xi}_{n_\xi}, y^{\xi}_{n_\xi})\right] \), different categorical levels can appear multiple times, once, or not at all. Although we initially sample \( n_\xi \) MMbedding vectors \(\left[\mathbf{b}^{\xi}_1, \dots, \mathbf{b}^{\xi}_i, \dots, \mathbf{b}^{\xi}_{n_\xi}\right]\), each sample corresponds to a specific level \( j \) of the categorical feature. To obtain the \(q\) MMbedding vectors per categorical level \(\left[ \mathbf{b}^{\xi}_1, \dots, \mathbf{b}^{\xi}_j, \dots, \mathbf{b}^{\xi}_q\right] \), we average all sampled vectors belonging to the same level, setting zeros for any levels absent in the current minibatch. This averaging procedure significantly reduces the number of parameters from \( q \times d \) per categorical feature, as employed by methods such as REbeddings, thereby mitigating overfitting risks. It is symbolized by \( \odot \) in Figure~\ref{fig:MMbeddings_scheme}, and is efficiently implemented in a vectorized manner to enhance computational performance. Furthermore, before computing the loss given by \eqref{eq:MMbeddings_loss}, we average the encoder outputs \( \pmb{\mu}^{\xi}_i \) and \( \log\left(\left[\pmb{\tau}^{\xi}_i\right]^2\right) \) across observations belonging to each categorical level \( j \):
\begin{equation*}
    \pmb{\mu}^{\xi}_j = \frac{1}{n_{\xi j}}\sum_{i \in \text{cluster } j}\pmb{\mu}^{\xi}_i, \quad \log\left(\left[\pmb{\tau}^{\xi}_j\right]^2\right) = \log\left[\frac{1}{n_{\xi j}^2}\sum_{i \in \text{cluster } j}\exp\left(\log\left(\left[\pmb{\tau}^{\xi}_i\right]^2\right)\right)\right],
\end{equation*}
where \( n_{\xi j} \) denotes the number of observations in cluster \( j \) within the minibatch \(\xi\), and \( \pmb{\mu}^{\xi}_j = \mathbf{0} \), \( \log\left(\left[\pmb{\tau}^{\xi}_j\right]^2\right) = \mathbf{0} \) if no observations from level \( j \) are present. While this averaging scheme may appear straightforward, it follows from a careful consideration of alternative strategies, as discussed in Appendix~\ref{app:averaging}.

Aggregating over the minibatch \(\xi\), and considering \( K \) categorical features each with cardinality \( q_k \) and embedding dimension \( d_k \), yields the full negative ELBO loss for MMbeddings:

\begin{equation}
\label{eq:MMbeddings_loss}
    \begin{aligned}
    -ELBO_\xi(f,q) &= \frac{n_\xi}{2}\log(2\pi\sigma^2) + \frac{1}{2\sigma^2} \sum_{i=1}^{n_\xi}\left(y^{\xi}_i - f\left(\mathbf{x}^{\xi}_i, \mathbf{b}^{\xi}_{(*1)}, \dots, \mathbf{b}^{\xi}_{(*K)}\right)\right)^2  \\
    & +\frac{\beta_V}{2} \sum_{k = 1}^K \sum_{j = 1}^{q_k}\sum_{m = 1}^{d_k}\left[-1 - \log\left(\left[\tau^{\xi}_{mjk}\right]^2\right) + \log\sigma^2_{mk} + \frac{\left(\mu^\xi_{mjk}\right)^2}{\sigma^2_{mk}} + \frac{\left(\tau^\xi_{mjk}\right)^2}{\sigma^2_{mk}}\right],
    \end{aligned}
\end{equation}
where \( \mathbf{b}^{\xi}_{(*1)}, \dots, \mathbf{b}^{\xi}_{(*K)} \) represent the corresponding sampled MMbeddings for each observation in the minibatch, and \( \beta_V \) is a hyperparameter borrowed from the \( \beta \)-VAE literature, which often enhances model performance \citep{betaVAE, Rolinek2018}. Note that the second term in the ELBO aggregates KL divergences over all categories and features, thus reflecting the full dataset, whereas the first term sums only over the minibatch; accordingly, balancing these terms requires scaling, and we empirically observe that setting \(\beta_V \approx \frac{n_\xi}{N}\) yields good results. The prior variances \(\sigma^2\) and \(\sigma^2_{mk}\) are necessary components of the model and can either be treated as fixed hyperparameters specified a priori, tuned through grid search, or optimized via SGD. In our implementation below, unless stated otherwise, we adopt the simplest prior assumption by fixing all prior variances to 1 during training and subsequently demonstrate the robustness of MMbeddings even when this simplifying assumption does not hold.

Typically, at inference time, our primary interest is in predicting outcomes for new, unseen data after training the encoder and decoder. Suppose \((\mathbf{X}_{tr}, \mathbf{y}_{tr})\) and \(\mathbf{X}_{te}\) denote the training and testing datasets, respectively. To reconstruct predictions \(\hat{\mathbf{y}}_{te}\), we first use the trained encoder to generate embedding vectors \(\mathbf{b}\) for all observations in the training set. These embeddings are grouped by their corresponding categorical levels, resulting in estimated MMbedding matrices \(\hat{\mathbf{B}}_k\) of size \(q_k \times d_k\) for each categorical feature \(k\). Finally, predictions for the testing data are obtained by applying the trained decoder \(\hat{f}\) to \(\mathbf{X}_{te}\) together with these embedding matrices, yielding \(\hat{\mathbf{y}}_{te} = \hat{f}(\mathbf{X}_{te}, \hat{\mathbf{B}}_1, \dots, \hat{\mathbf{B}}_K)\). In practice, we often found it beneficial to freeze the embedding matrices \(\hat{\mathbf{B}}_k\) after initial training and subsequently fine-tune only the decoder for several additional epochs.

\section{Experiments}
\label{sec:experiments}
We now present experimental results using both simulated and real datasets to demonstrate the effectiveness of the MMbeddings approach. We begin our analysis with simulated data, designed specifically to capture scenarios involving high-cardinality categorical features. All experiments were conducted using Python in Google Colab with Keras on TensorFlow, and code and data are publicly available at \url{https://github.com/gsimchoni/mmbeddings}.

\subsection{Simulated data}
\label{ssec:simulated}
We generate data according to model \eqref{eq:MMbeddings_scalar}, where each observation \(\mathbf{x}_{ij}\) contains \(p = 10\) features sampled uniformly from \(\mathcal{U}(-1, 1)\). The embeddings \(\mathbf{b}_j\) are \(d = 10\)-dimensional vectors drawn from \(\mathcal{N}(\mathbf{0}, \mathbf{D})\), with covariance matrix \(\mathbf{D} = \sigma_b^2 \mathbf{I}_d\), and variance parameter \(\sigma_b^2\) fixed on 1, or varied in further experiments in Appendix~\ref{app:simulated}. The nonlinear function \(f\), inspired by Lindstrom and Bates, uniquely combines each feature with fixed and random effects from the \(\mathbf{B}_{q \times d}\) embeddings and can be viewed in Appendix~\ref{app:f_formula}.

Additionally, a noise term \(\varepsilon_{ij}\) is introduced from \(\mathcal{N}(0, \sigma^2)\) with \(\sigma^2 = 1\), and this \(f(\mathbf{X}, \mathbf{B}) + \pmb{\varepsilon}\) is used as \(\mathbf{y}\) for regression or binarized for classification. We examine scenarios with increasing cardinality \(q \in \{10^2, 10^3, 10^4, 10^5\}\), maintaining \(n = 10q\) overall, with varying group sizes \(n_j\), sampled from a multinomial distribution.

We compare MMbeddings against alternative embedding methods, including standard embeddings, L2-regularized embeddings (EmbedL2), REbeddings, UEmbeddings, and a baseline model which is agnostic to the categorical feature (using only the feature matrix \(\mathbf{X}\)). Other methods such as mean-encoding, PCA-encoding, feature hashing, LMMNN, TabNet and TabTransformer were also tested but performed badly, and their detailed outcomes are provided in Appendix~\ref{app:simulated}. Each method employs a decoder architecture with a two-layer MLP with two layers of 10 neurons each, and ReLU activations. MMbeddings adds an encoder which consists of a two-layer MLP with layer sizes of 100 neurons each, also using ReLU activations. All models were trained using early stopping based on performance on a 10\% validation set, with patience set to 10 epochs and batch size equal to \(q\). Training proceeded for a maximum of 1000 epochs. For MMbeddings specifically, the parameter \(\beta_V\) was fixed at 0.001, as the performance was stable across similarly small values. The latent dimension \(d\) was treated as known in all cases, with sensitivity analyses of this parameter included in Appendix~\ref{app:simulated}.

Each experiment was repeated 10 times, evaluating performance on 100K unseen test observations. For regression tasks, we measured mean squared error \(MSE_Y\); for classification, we used the area under the ROC curve \(AUC_Y\). Embedding quality was assessed by computing the normalized root mean squared error (\(RMSE_D\)) between pairwise Euclidean \(q \times q\) distances of true embeddings \(\mathbf{B}\) and estimated embeddings \(\hat{\mathbf{B}}\) (see Appendix \ref{app:rmse_d} for details). Recognizing the practical significance of embeddings for downstream tasks, we also simulated a binary outcome from a linear model using the true embeddings and reported the resulting \(AUC_B\) after fitting logistic regression on the estimated embeddings \(\hat{\mathbf{B}}\).

Table~\ref{tab:results:simulated} shows the performance comparison across embedding methods for simulated regression and classification tasks with varying cardinalities. MMbeddings consistently outperforms alternative methods in predictive performance (\(MSE_Y\) and \(AUC_Y\)), embedding quality (\(RMSE_D\)), and downstream utility (\(AUC_B\)). Improvements become particularly pronounced with increasing cardinality, underscoring MMbeddings' robustness in handling high-cardinality categorical features. Notably, MMbeddings maintains a fixed and relatively small number of parameters (13,651 parameters for all cases above) independent of the cardinality \(q\). In contrast, traditional embedding methods require significantly more parameters -- linear in \(q\) -- making MMbeddings substantially less prone to overfitting, especially for larger \(q\). Additional experiments examining inclusion of multiple categorical features, and other variations in simulation parameters are detailed in Appendix~\ref{app:simulated}, including scatterplots of predicted versus true embeddings for each latent dimension (Figure~\ref{fig:sim_b_pred_vs_true}).
 
\begin{table}
    \centering
\begin{tabular}{ll|llllll}
\hline
                            &              & Ignore     & Embed. & EmbedL2 & REmbed. & UEmbed. & MMbed. \\
\hline
$q$               & Metric       & \multicolumn{6}{c}{Regression}                                                     \\
\hline
\multirow{3}{*}{$10^2$}  & $MSE_{Y}$        & 5.29 (.18) & 3.67 (.11) & 3.74 (.14) & 5.25 (.17) & 4.09 (.11) & \textbf{3.46 (.11)} \\
                            & $RMSE_{D}$    & -- & 0.47 (.01) & 0.50 (.01) & 0.48 (.01) & 0.47 (.01) & \textbf{0.29 (.01)} \\
                            & $AUC_{B}$ & -- & 0.68 (.01) & 0.65 (.02) & 0.64 (.01) & 0.66 (.02) & \textbf{0.71 (.01)} \\
\hline
\multirow{3}{*}{$10^3$}   & $MSE_{Y}$        & 4.90 (.35) & 3.98 (.28) & 3.95 (.23) & 4.82 (.36) & 4.06 (.10) & \textbf{3.36 (.20)} \\
                            & $RMSE_{D}$    & -- & 0.40 (.01) & 0.43 (.01) & 0.40 (.01) & 0.41 (.01) & \textbf{0.27 (.01)} \\
                            & $AUC_{B}$ & -- & 0.68 (.01) & 0.63 (.02) & 0.67 (.02) & 0.67 (.01) & \textbf{0.71 (.02)} \\
\hline
\multirow{3}{*}{$10^4$}  & $MSE_{Y}$        & 5.12 (.29) & 3.42 (.12) & 3.51 (.16) & 3.92 (.17) & 3.95 (.11) & \textbf{3.01 (.09)} \\
                            & $RMSE_{D}$    & -- & 0.34 (.01) & 0.37 (.01) & 0.27 (.01) & 0.36 (.01) & \textbf{0.23 (.01)} \\
                            & $AUC_{B}$ & -- & 0.69 (.03) & 0.62 (.04) & \textbf{0.72 (.03)} & 0.69 (.01) & \textbf{0.73 (.03)} \\
\hline
\multirow{3}{*}{$10^5$} & $MSE_{Y}$        & 4.93 (.21) & 3.60 (.11) & 3.70 (.11) & 3.50 (.10) & 3.95 (.11) & \textbf{3.10 (.09)} \\
                            & $RMSE_{D}$    & -- & 0.35 (.01) & 0.38 (.01) & \textbf{0.21 (.01)} & 0.34 (.01) & 0.24 (.01) \\
                            & $AUC_{B}$ & -- & 0.69 (.02) & 0.58 (.02) & \textbf{0.73 (.02)} & 0.70 (.01) & \textbf{0.72 (.01)} \\
\hline
                            &              & \multicolumn{6}{c}{Classification}                                                 \\
\hline
\multirow{3}{*}{$10^2$}  & $AUC_{Y}$        & 0.67 (.01) & \textbf{0.82 (.01)} & \textbf{0.80 (.01)} & 0.67 (.01) & 0.78 (.01) & \textbf{0.82 (.01)} \\
                            & $RMSE_{D}$    & -- & 0.48 (.01) & 0.52 (.01) & 0.49 (.01) & 0.48 (.02) & \textbf{0.33 (.01)} \\
                            & $AUC_{B}$ & -- & 0.66 (.02) & 0.63 (.02) & 0.67 (.02) & 0.61 (.01) & \textbf{0.72 (.01)} \\
\hline
\multirow{3}{*}{$10^3$}   & $AUC_{Y}$        & 0.71 (.01) & 0.82 (.01) & 0.81 (.01) & 0.74 (.01) & 0.78 (.00) & \textbf{0.84 (.00)} \\
                            & $RMSE_{D}$    & -- & 0.40 (.01) & 0.42 (.01) & 0.37 (.01) & 0.43 (.01) & \textbf{0.26 (.01)} \\
                            & $AUC_{B}$ & -- & 0.62 (.01) & 0.58 (.01) & \textbf{0.65 (.01)} & 0.63 (.01) & \textbf{0.65 (.01)} \\
\hline
\multirow{3}{*}{$10^4$}  & $AUC_{Y}$        & 0.71 (.01) & 0.82 (.01) & 0.82 (.01) & 0.78 (.01) & 0.79 (.01) & \textbf{0.85 (.01)} \\
                            & $RMSE_{D}$    & -- & 0.36 (.00) & 0.38 (.00) & 0.31 (.01) & 0.36 (.01) & \textbf{0.24 (.00)} \\
                            & $AUC_{B}$ & -- & 0.65 (.01) & 0.61 (.02) & \textbf{0.68 (.01)} & \textbf{0.69 (.01)} & \textbf{0.69 (.01)} \\
\hline
\multirow{3}{*}{$10^5$} & $AUC_{Y}$       & 0.71 (.01) & 0.83 (.01) & 0.71 (.01) & 0.81 (.01) & 0.79 (.01) & \textbf{0.86 (.01)} \\
                            & $RMSE_{D}$    & -- & 0.35 (.01) & 0.38 (.01) & 0.27 (.01) & 0.36 (.01) & \textbf{0.25 (.01)} \\
                            & $AUC_{B}$ & -- & 0.67 (.02) & 0.50 (.00) & 0.\textbf{70 (.02)} & 0.68 (.02) & \textbf{0.71 (.02)} \\
\hline
\end{tabular}

    \caption{Performance metrics across embedding methods on simulated datasets for regression and classification tasks with a single high-cardinality categorical feature of cardinality \(q\). Mean test predictive performance (\(MSE_Y\) and \(AUC_Y\)), embedding quality (\(RMSE_D\)), and downstream embedding utility (\(AUC_B\)) are reported, with standard errors in parentheses. Bold indicates non-inferior to the best performance per row by a paired t-test.}
    \label{tab:results:simulated}
\end{table}

\begin{figure}
    \centering
    \includegraphics[width=0.85\linewidth]{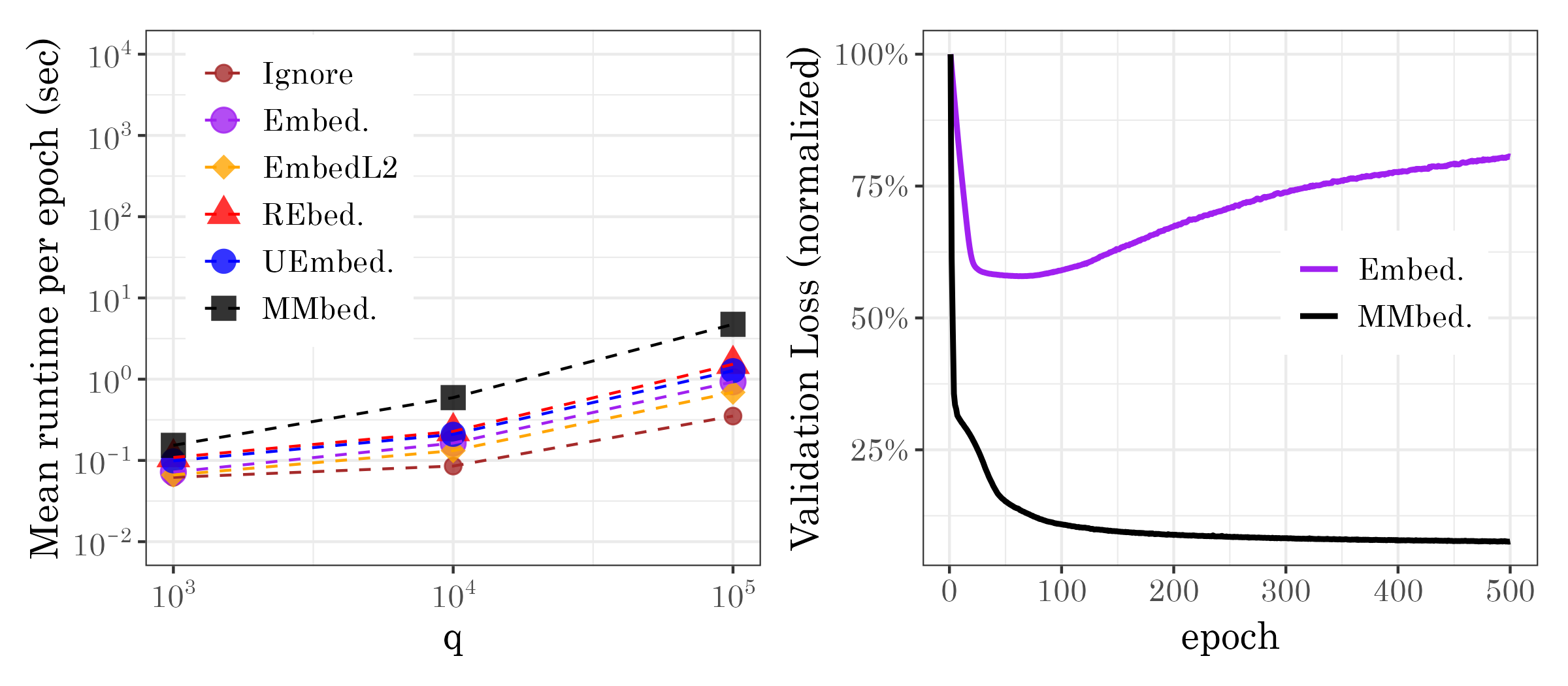}
    \caption{Left: Mean runtime per epoch (log-log scale) is shown for varying cardinality \(q\) (with sample size \(n = 10q\)), regression setting. Right: The validation loss plot (normalized) for \(q = 10^4\).}
    \label{fig:results:increase_q_val_loss}
\end{figure}

To conclude this part, Figure~\ref{fig:results:increase_q_val_loss} compares computational efficiency and generalization performance of MMbeddings with competing methods on the simulated regression data. The left panel shows mean runtime per epoch against increasing categorical cardinality \(q\), with the sample size \(n\) set to \(10q\). While MMbeddings scale similarly to alternative approaches, their computational overhead is higher compared to standard embeddings. The right panel illustrates validation loss over epochs for \(q = 10^4\). Standard embeddings quickly exhibit severe overfitting, indicating that early stopping or additional regularization is essential. In contrast, MMbeddings substantially reduce overfitting while using roughly one-tenth of the embedding parameters.

\subsection{Real data}
\label{ssec:real}
We now present results for MMbeddings applied to real-world datasets. We demonstrate our method's capabilities first on a substantial classification dataset within a collaborative filtering framework, and subsequently on several regression datasets characterized by high-cardinality categorical features.

\paragraph{Classification with Neural Collaborative Filtering}
We test MMbeddings within the neural collaborative filtering (NCF) framework \citep{CollaborativeFiltering, NeuralCF}, using the Amazon video games ratings dataset \citep{ni2019justifying}. The dataset originally consists of approximately 1.3 million tuples containing user IDs, video game IDs, ratings, and timestamps. We sampled 100K user-game interactions with ratings of 5, interpreting these as positive interactions. Following the implicit feedback methodology of \citep{NeuralCF}, we consider these interactions as positive samples (labeled "1") and additionally generate 100K negative samples (labeled "0") by selecting user-game combinations not present in the original data. Timestamps for negative samples were drawn from each user's existing timestamp distribution. Hence, our task transforms into a challenging binary classification problem involving recommending items (games) to users, with additional temporal features. Our dataset thus comprises \(n = 200K\) total samples, \(p = 10\) continuous timestamp-based features, and two high-cardinality categorical features representing users (\(q_{users} = 90K\)) and games (\(q_{games} = 20K\)).

We implement the NCF architecture by embedding users and items into latent vectors, calculating their inner products, concatenating additional temporal features, and then applying a decoder to output probabilities using a sigmoid activation. Specifically, we compare standard embedding layers and UEmbeddings versus our MMbeddings method, integrating the variational loss component. For MMbeddings, we use a single-layer encoder of 100 neurons and ReLU activation, all variance priors were set to 0.5, and the variational hyperparameter \(\beta_V\) was fixed at 0.1. All models were trained using 10-fold cross-validation with a batch size varying in \(\{10^3, 10^4\}\), an early stopping procedure with \(n_p \in \{2, 5, 10\}\) epochs patience monitoring a 10\% validation set, with a latent dimension \(d = 10\) and the same decoder -- a two-layer MLP with 10 neurons in each layer. UEmbeddings performed very poorly across multiple choices of \(q_{U}\); thus, we omit their results here. This outcome is understandable, as the inner product-based NCF architecture does not naturally align with the unified embedding framework of UEmbeddings.

Table~\ref{tab:results:real:amazon} shows mean AUC, logloss, and accuracy on unseen data. MMbeddings consistently outperform standard embeddings with about two orders of magnitude fewer parameters (5.4K vs. 1.2M). While standard embeddings quickly overfit (increasing logloss as \(n_p\) grows), MMbeddings maintain stable or slightly improved performance, highlighting robustness in high-cardinality scenarios.

\begin{table}
    \centering
    \begin{tabular}{ll|lll|lll}
\hline
                          &       & \multicolumn{3}{l|}{NCF-Embeddings (1.2M params)}   & \multicolumn{3}{l}{NCF-MMbeddings (5.4K params)}   \\
                          \hline
                          & batch & $n_p = 2$  & $n_p = 5$  & $n_p = 10$ & $n_p = 2$  & $n_p = 5$  & $n_p = 10$ \\
                          \hline
\multirow{2}{*}{AUC}      & $10^3$  & .69 (.006) & .65 (.001) & .65 (.002) & .69 (.001) & .70 (.001) & .70 (.001) \\
                          & $10^4$ & .68 (.004) & .66 (.006) & .65 (.003) & .70 (.001) & .70 (.001) & .71 (.001) \\
                          \hline
\multirow{2}{*}{LogLoss}  & $10^3$ & .84 (.021) & 1.6 (.016) & 2.2 (.016) & .64 (.000) & .64 (.001) & .64 (.001) \\
                          & $10^4$ & .66 (.005) & .77 (.013) & 1.0 (.028) & .64 (.002) & .64 (.002) & .64 (.001) \\
                          \hline
\multirow{2}{*}{Accuracy} & $10^3$ & .64 (.004) & .60 (.002) & .60 (.001) & .65 (.001) & .65 (.001) & .65 (.001) \\
                          & $10^4$ & .65 (.003) & .63 (.006) & .61 (.002) & .65 (.001) & .65 (.001) & .66 (.001) \\
\hline
\end{tabular}

    \caption{Performance of standard embeddings vs. MMbeddings in NCF on the Amazon video games ratings dataset, across varying batch sizes and patience epochs (\(n_p\)), with 10-CV standard errors in parentheses. MMbeddings' performance improvements (AUC, accuracy), and reductions (LogLoss), are statistically significant under a paired t-test.}
    \label{tab:results:real:amazon}
\end{table}

\paragraph{Regression with TabTransformer}
For regression, we evaluate MMbeddings on the five datasets from \citep{lmmnn_jmlr}, each containing multiple high-cardinality categorical features (see Table~\ref{tab:results:real:regression} for a concise overview and Appendix~\ref{app:datasets} for more extensive descriptions). As an illustrative example, the UK Biobank (UKB) dataset \citep{UKB} comprises 42K cancer patients with five high-cardinality categorical variables: diagnosis, operation, treatment, cancer type, and tumor histology. The task involves predicting standardized triglyceride levels in the blood (mmol/L).

We compare standard embeddings, UEmbeddings and MMbeddings using the TabTransformer architecture, effective for tabular data. MMbeddings employs a single-layer encoder (100 neurons, ReLU) outputting means and log-variances, with fixed \(\beta_V = 0.1\) and variance priors at 0.1. All methods share latent dimension \(d = 10\), early stopping (patience = 5 epochs, 10\% validation), and an identical decoder (two layers, 10 neurons each) on the transformer output. Batch size is optimized per method and \(q_{U}\) for UEmbeddings.

Critically, although all methods share identical inference-time architectures, standard embeddings require substantially more parameters, proportional to each categorical cardinality \(q_k\), resulting in 3-70 times greater complexity and potential overfitting. Table~\ref{tab:results:real:regression} summarizes results, demonstrating MMbeddings consistently achieve superior predictive performance with fewer parameters.

\begin{table}
    \centering
    \begin{tabular}{l|lll|l|lll}
\hline
    Dataset & $n$, $p$ & categorical ($q$) & $y$ & Metric & TT-Em. & TT-UEm. & TT-MM.\\
    \hline
    \multirow{2}{*}{}
    Imdb   & 86K, 159 & director (38K) & movie & MSE & 1.20 (.02) & 1.29 (.01) & \textbf{1.14 (.01)} \\
           &   & movie type (2K) & ratings & param & 407K & 19K & 29K\\
    \hline
    \multirow{2}{*}{}
    News   & 81K, 176 & source (5.4K) & item & MSE & 2.57 (.01) & 2.45 (.02) & \textbf{2.36 (.02)}\\
           &   & title (72K) & shares & param & 790K & 90K & 11K \\
    \hline
    \multirow{3}{*}{}
    IE   & 73K, 3 & student (2.9K) & teacher & MSE & 1.51 (.01) & 1.51 (.01) & \textbf{1.49 (.00)} \\
               &   & teacher (1.1K) & ratings & param & 48K & 88K & 13K \\
               &   & department (14) & & & \\
    \hline
    \multirow{4}{*}{}
    Spotify   & 28K, 14 & artist (10K) & song & MSE & .014 (.00) & \textbf{.012 (.00)} & \textbf{.012 (.00)} \\
           &   & album (22K) & dance- & param & 368K & 17K & 17K \\
           &   & playlist (2.3K) & ability & & \\
           &   & subgenre (553) & & & \\
    \hline
    \multirow{5}{*}{}
    UKB   & 42K, 19 & treatment (1.1K) & triglyc. & MSE & 0.99 (.02) & \textbf{0.95 (.01)} & \textbf{0.94 (.01)} \\
           &   & operation (2.0K) & level & param & 68K & 17K & 19K \\
           &   & diagnosis (2.1K) & & & \\
           &   & cancer type (446) & & & \\
           &   & histology (359) & & & \\
    \hline
 \end{tabular}
    \caption{Performance of standard embeddings, UEmbeddings and MMbeddings in TabTransformer on \citep{lmmnn_jmlr}'s categorical regression datasets, with 10-CV standard errors in parentheses. Bold indicates non-inferior to the best performance per row by a paired t-test.}
    \label{tab:results:real:regression}
\end{table}

\section{Conclusion}
\label{sec:conclusion}
We introduced MMbeddings, a novel method that integrates random effects from NLMM into a VAE to model categorical embeddings. This approach reduces parameter complexity and mitigates overfitting in high-cardinality settings. Experiments on simulated and real data show that MMbeddings outperform traditional embeddings and other encoding methods in both accuracy and robustness.

Our method notably maintains a fixed and modest parameter count, independent of the categorical feature cardinality. This robustness becomes particularly pronounced as cardinality increases, positioning MMbeddings as a practical and theoretically grounded solution for modern predictive modeling challenges involving categorical data.

For future research, MMbeddings can naturally incorporate more complex, non-diagonal covariance structures when additional relational information, such as knowledge graphs between categorical levels, is available. Although our current diagonal covariance assumption simplifies computations and has empirically demonstrated strong performance, leveraging known relationships between categories -- for instance, the number of shared movies between movie raters \citep{SEE_neurips2019} -- could further improve embedding quality and predictive capabilities.

Additionally, preliminary results indicate promising extensions of MMbeddings to sequence and text data, where embeddings typically take the form of three-dimensional tensors \((n_b, n_{\text{sequence}}, d)\). We have working code for such settings, demonstrating good initial performance, suggesting the potential broad applicability and scalability of our approach beyond tabular data.

\begin{ack}
This study was supported in part by a fellowship from the Edmond J. Safra Center for Bioinformatics at Tel-Aviv University, by Israel Science Foundation grant 2180/20 and by Israel Council for Higher Education Data-Science Centers.
\end{ack}

\newpage
\bibliographystyle{plainnat}
\bibliography{references}


\appendix

\section{Combining the means and variances for each cluster}
\label{app:averaging}
In constructing the MMbeddings framework, a natural question arises regarding how best to aggregate encoder outputs for observations belonging to the same categorical level \( j \) within a minibatch \( \xi \). Our current implementation computes the average of both the means \( \pmb{\mu}^{\xi}_i \) and the (log) variances \( \log\left(\left[\pmb{\tau}^{\xi}_i\right]^2\right) \) across all observations assigned to level \( j \). While this approach may seem like a straightforward design choice, it stems from a deeper analysis of the statistical structure underlying RE models.

To motivate this choice, consider the posterior distribution of a random intercept \( b_{0j} \) in a standard LMM:
\[
y_{ij} = \mathbf{x}_{ij}^T\pmb{\beta} + b_{0j} + \varepsilon_{ij}, \quad b_{0j} \sim \mathcal{N}(0, \sigma^2_b), \quad \varepsilon_{ij} \sim \mathcal{N}(0, \sigma^2).
\]
The posterior mean of \( b_{0j} \), given the \( n_j \) observations from cluster \( j \), is well known to be a shrinkage estimator:
\[
\mathbb{E}[b_{0j} \mid \mathbf{X}_j, \mathbf{y}_j] = \frac{\sigma^2_b}{\sigma^2/n_j + \sigma^2_b}( \bar{y}_j - \bar{\mathbf{x}}_j^T\pmb{\beta}).
\]
Assuming \( \sigma^2 = \sigma^2_b \), this expression simplifies to
\[
\frac{n_j}{n_j + 1}( \bar{y}_j - \bar{\mathbf{x}}_j^T\pmb{\beta}).
\]
This offers an intuitive interpretation: the posterior mean interpolates between the prior (zero) and the within-cluster residual mean. When \( n_j = 1 \), the shrinkage factor becomes \( \frac{1}{2} \); as \( n_j \to \infty \), it approaches 1, recovering the full residual mean.

Likewise, the posterior variance of \( b_{0j} \) is given by:
\[
\mathrm{Var}[b_{0j} \mid \mathbf{X}_j, \mathbf{y}_j] = \left( \frac{n_j}{\sigma^2} + \frac{1}{\sigma^2_b} \right)^{-1},
\]
which simplifies under the same assumption to \( \frac{\sigma^2}{n_j + 1} \), again shrinking to 0 as \( n_j \to \infty \).

These classical results suggest that the degree of shrinkage -- both in mean and variance -- should depend on the number of observations per cluster. In our setting, however, the encoder does not explicitly account for the number of observations per categorical level when generating its outputs \( \pmb{\mu}_i \) and \( \pmb{\tau}_i \). This led us to explore whether applying a corrective shrinkage factor \( c_j \) to the averaged encoder outputs could improve performance. Specifically, we considered two strategies:
\begin{itemize}
    \item \textbf{Heuristic shrinkage:} Multiplying the averaged outputs for cluster \( j \) by a fixed factor \( c_j = \frac{n_{\xi,j}}{n_{\xi,j} + 1} \), mimicking the classical LMM expression.
    \item \textbf{Learned scaling:} Introducing a learnable parameter \( c_j \) for each cluster, trained via backpropagation.
\end{itemize}

Surprisingly, empirical comparisons on both simulated and real datasets revealed that the best-performing strategy was to use simple averaging with no additional scaling. This was somewhat unexpected, given that in many practical scenarios -- including high-cardinality categorical data -- the average minibatch contains at most one representative per level, i.e., \( n_{\xi,j} \in \{0,1\} \) for most \( j \). The result might suggest that attempts to impose additional shrinkage introduce more noise than benefit when \( n_{\xi,j} \) is small and unstable.

In summary, although richer schemes inspired by LMM theory are plausible and theoretically grounded, the naive averaging approach seems to provide an effective empirical solution across sparse and dense cluster regimes alike.

\section{Formula for the non-linear function used in simulated data}
\label{app:f_formula}
\[
\begin{aligned}
    f(\mathbf{X}, \mathbf{B}) &= 0.1 e^{-\beta_{1j} X_{1}^2}
+ \text{clip}\left(\frac{\beta_{2j} X_{2}}{1 + \beta_{3j} X_{3}^2}, -5, 5\right)
+ \sin(\beta_{4j} X_{4})
+ \beta_{5j} X_{5} \\
&+ \frac{\beta_{6j} X_{6}}{1 + e^{-\beta_{7j} X_{7}}}
+ \arctan(\beta_{8j} X_{8})
+ \beta_{9j}\cos(X_{9})
+ \beta_{10j}\log(1 + X_{10}^2),
\end{aligned}
\]

where \(\beta_{mj} = \beta_m + B_{jm}\) for \(m = 1, \dots, 10\), and \(\beta_m = 1\) for all \(m\).

\section{A formula for normalized RMSE between distance matrices}
\label{app:rmse_d}
\[
\text{RMSE}_D = \frac{\sqrt{\frac{1}{\frac{q(q-1)}{2}} \sum_{i<j}(D_{true}^{ij}-D_{est}^{ij})^2}}{\max(D_{true}) - \min(D_{true})},
\]
where \(D_{true}\) and \(D_{est}\) are \(q \times q\) distance matrices of the true and estimated embeddings, respectively.

\section{Simulated datasets additional results}
\label{app:simulated}

\begin{table}[H]
    \centering
\begin{tabular}{l|llllll}
\hline
                            & Ignore     & Embed. & EmbedL2 & REmbed. & UEmbed. & MMbed. \\
\hline
Metric       & \multicolumn{6}{c}{\(\sigma^2_m = 0.3\)} \\
\hline
$MSE_{Y}$        & 1.71 (.04) & 1.70 (.02) & \textbf{1.57 (.02)} & 1.72 (.03) & 1.73 (.02) & \textbf{1.63 (.02)} \\
                            $RMSE_{D}$    & -- & 0.32 (.01) & 0.38 (.00) & 0.29 (.01) & 0.33 (.02) & \textbf{0.16 (.00)} \\
                            $AUC_{B}$ & -- & 0.61 (.02) & 0.57 (.02) & 0.63 (.02) & 0.60 (.02) & \textbf{0.65 (.02)} \\
\hline
 & \multicolumn{6}{c}{\(\sigma^2_m = 3.0\)} \\
\hline
$MSE_{Y}$        & 8.84 (.59) & 5.47 (.16) & 5.99 (.24) & 6.58 (.27) & 7.11 (.21) & \textbf{4.88 (.14)} \\
                            $RMSE_{D}$    & -- & 0.37 (.00) & 0.39 (.00) & 0.31 (.01) & 0.39 (.01) & \textbf{0.26 (.01)} \\
                            $AUC_{B}$ & -- & \textbf{0.64 (.02)} & 0.57 (.03) & \textbf{0.65 (.02)} & 0.63 (.01) & \textbf{0.65 (.02)} \\
\hline
\end{tabular}

    \caption{Performance metrics across embedding methods on simulated datasets for regression with \(q = 10^4\), under varying true priors. Bold indicates non-inferior to the best performance per row by a paired t-test.}
    \label{tab:diff_sig2bs}
\end{table}

\begin{table}[ht]
    \centering
\begin{tabular}{ll|lll|lll}
\hline
                            \textbf{$q$}               & Metric & $\sigma^2_m = 0.1$     & $\sigma^2_m = 1.0$ (*) & $\sigma^2_m = 2.0$ & $d = 5$ & $d = 10$ (*) & $d = 20$ \\
\hline
\multirow{3}{*}{$10^3$}  & $MSE_{Y}$        & 3.22 (.09) & 3.34 (.07) & 3.09 (.07) & 3.22 (.09) & 3.34 (.07) & 3.21 (.08) \\
                            & $RMSE_{D}$    & 0.34 (.01) & 0.26 (.01) & 0.24 (.01) & 0.27 (.01) & 0.26 (.01) & 0.25 (.01) \\
                            & $AUC_{B}$ & 0.70 (.02) & 0.71 (.01) & 0.71 (.02) & 0.72 (.02) & 0.71 (.01) & 0.71 (.02) \\
\hline
\multirow{3}{*}{$10^4$}   & $MSE_{Y}$        & 3.06 (.08) & 2.95 (.07) & 2.94 (.06) & 3.21 (.09) & 2.95 (.07) & 3.03 (.08) \\
                            & $RMSE_{D}$    & 0.31 (.00) & 0.24 (.00) & 0.21 (.00) & 0.25 (.00) & 0.24 (.00) & 0.22 (.00) \\
                            & $AUC_{B}$ & 0.69 (.01) & 0.69 (.01) & 0.71 (.02) & 0.69 (.02) & 0.69 (.01) & 0.71 (.02) \\
\hline
\end{tabular}

    \caption{Performance of MMbeddings under incorrect hyperparameter specifications in simulated regression scenarios (true prior variance \(\sigma^2_m = 1\), true latent dimension \(d = 10\)). Mean predictive performance (\(MSE_Y\)), embedding quality (\(RMSE_D\)), and downstream embedding utility (\(AUC_B\)) are reported with standard errors in parentheses, varying either the prior variance or latent dimension. Interestingly, for \(q = 10^3\), the true parameter values (*) do not yield optimal performance.}
    \label{tab:varying_reprior_and_d}
\end{table}

\begin{table}[ht]
    \centering
\begin{tabular}{l|llllll}
\hline
                            & LMMNN     & TabNet(Embed.) & TabTrans.(Embed.) & FeatHash. & MeanEnc & MMbed. \\
\hline
$q$ & \multicolumn{5}{c}{Regression \(MSE_Y\)}                                                     \\
\hline
$10^2$ & 3.84 (.10) & 4.99 (.11) & 3.98 (.10) & 3.60 (.12) & 5.14 (.17) & \textbf{3.46 (.11)} \\
$10^3$ & \textbf{3.38 (.11)} & 5.02 (.24) & 4.23 (.21) & 4.20 (.28) & 4.90 (.36) & \textbf{3.36 (.20)} \\
$10^4$  & 3.85 (.35) & 5.32 (.31) & 3.93 (.12) & 4.97 (.28) & 5.13 (.30) & \textbf{3.01 (.09)} \\
$10^5$ & 3.51 (.10) & 4.42 (.16) & 4.10 (.10) & 4.94 (.20) & 4.93 (.21) & \textbf{3.10 (.09)} \\
\hline
                            & \multicolumn{5}{c}{Classification \(AUC_Y\)}                                                 \\
\hline
$10^2$ & \textbf{0.82 (.01)} & 0.65 (.01) & 0.80 (.01) & \textbf{0.82 (.01)} & 0.70 (.01) & \textbf{0.82 (.01)} \\
$10^3$ & 0.82 (.02) & 0.67 (.01) & 0.80 (.01) & 0.79 (.00) & 0.71 (.01) & \textbf{0.84 (.00)} \\
$10^4$ & 0.81 (.02) & 0.67 (.00) & 0.80 (.01) & 0.72 (.00) & 0.71 (.01) & \textbf{0.85 (.01)} \\
$10^5$ & 0.83 (.01) & 0.78 (.01) & 0.81 (.01) & 0.71 (.01) & 0.71 (.01) & \textbf{0.86 (.01)} \\
\hline
\end{tabular}

    \caption{Performance metrics for more architectures and encodings in addition to Table~\ref{tab:results:simulated} on simulated datasets for regression and classification tasks with a single high-cardinality categorical feature of cardinality \(q\). Mean test predictive performance (\(MSE_Y\) and \(AUC_Y\)) are reported, with standard errors in parentheses. Bold indicates non-inferior to the best performance per row by a paired t-test.}
    \label{tab:more_methods_single_feature}
\end{table}

\begin{table}[ht]
    \centering
\begin{tabular}{l|llll}
\hline
 & \multicolumn{4}{c}{Regression}                                                     \\
\hline
                            Method & \(MSE_Y\)     & \(RMSE_D\) & \(AUC_D\) & \(n_{params}\)\\
\hline
MMbed. & \textbf{11.19 (.21)} & \textbf{0.28 (.00)} & \textbf{0.57 (.01)} & 17891 \\
Embed. & 12.32 (.27) & 0.36 (.00) & \textbf{0.59 (.01)} & 300531 \\
EmbedL2 & 11.47 (.25) & 0.37 (.00) & \textbf{0.59 (.01)} & 300531 \\
REmbed. & 12.24 (.31) & 0.32 (.00) & \textbf{0.60 (.01)} & 600531 \\
UEmbed. & 12.36 (.27) & -- & -- & 10486291 \\
FeatHash. & 13.28 (.36) & -- & -- & 30951 \\
LMMNN & 11.91 (.19) & -- & -- & 235 \\
TabNet & 19.46 (.59) & -- & -- & 339779 \\
TabTrans. & 20.52 (.39) & -- & -- & 306829 \\
MeanEnc & 12.88 (.38) & -- & -- & 531 \\
Ignore & 12.82 (.39) & -- & -- & 231 \\
\hline
 & \multicolumn{4}{c}{Classification}                                                     \\
\hline
                            Method & \(AUC_Y\)     & \(RMSE_D\) & \(AUC_D\) & \(n_{params}\)\\
\hline
MMbed. & \textbf{0.82 (.00)} & \textbf{0.28 (.00)} & 0.55 (0.01) & 17891 \\
Embed. & 0.80 (.00) & 0.36 (.00) & \textbf{0.59 (0.01)} & 300531 \\
EmbedL2 & 0.81 (.00) & 0.37 (.00) & 0.50 (.00) & 300531 \\
REmbed. & 0.79 (.00) & 0.36 (.00) & \textbf{0.60 (0.01)} & 600531 \\
UEmbed. & 0.80 (.00) & -- & -- & 10486291 \\
FeatHash. & 0.79 (.00) & -- & -- & 30951 \\
LMMNN & -- & -- & -- & -- \\
TabNet & 0.74 (.01) & -- & -- & 339779 \\
TabTrans. & 0.68 (.01) & -- & -- & 306829 \\
MeanEnc & 0.80 (.00) & -- & -- & 531 \\
Ignore & 0.80 (.00) & -- & -- & 231 \\
\hline
\end{tabular}

    \caption{Performance metrics on simulated datasets with three high-cardinality categorical features of cardinality \([10^4, 10^4, 10^4]\) and \(n = 100K\). The same architecture as in the single feature experiment is used, with a batch size of \(10^4\). Mean test predictive performance (\(MSE_Y\)), embedding quality (\(RMSE_D\)), and downstream embedding utility (\(AUC_B\)) are reported, with standard errors in parentheses. Bold indicates non-inferior to the best performance per row by a paired t-test. Notice LMMNN cannot currently fit multiple categorical features for classification.}
    \label{tab:more_methods_multiple_features}
\end{table}

\begin{figure}
    \centering
    \includegraphics[width=1.0\linewidth]{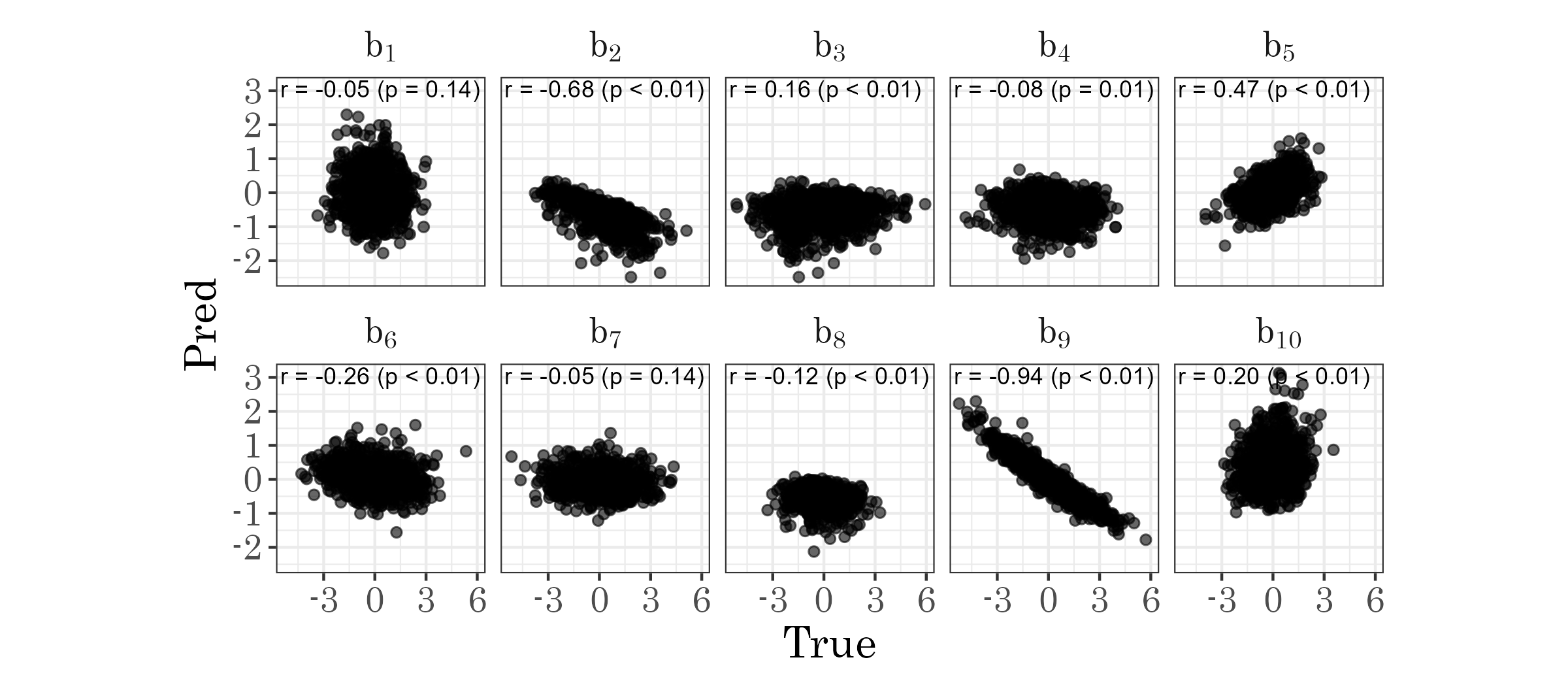}
    \caption{Predicted vs. true embedding values for each component \(\mathbf{b}_1\) to \(\mathbf{b}_{10}\), for the \(q = 10^3\) cardinality scenario, regression setting. For each true component (\(\mathbf{B}\)), the most correlated predicted component (column in (\(\hat{\mathbf{B}}\)) was selected based on Pearson correlation. Axis limits are shared across facets to ensure comparability.}
    \label{fig:sim_b_pred_vs_true}
\end{figure}

\section{Real datasets additional details}
\label{app:datasets}
\begin{table}[H]
    \centering
    \begin{tabular}{p{0.1\linewidth}p{0.15\linewidth}p{0.11\linewidth}p{0.15\linewidth}p{0.35\linewidth}}
    \hline
    Dataset & Source & Availability & Reference & Description \\
    \hline
    Imdb & Kaggle & Free & \citet{wrandrall_2021} & 86K movie titles scraped from imdb.com along with their genre, director, date of release a 1-10 mean score and a textual description which is processed to top 1-gram tokens count. \\
    News & UCI ML & Free & \citet{Moniz2018} & 81K news items and their number of shares on Facebook. Headline is processed to top 1-gram tokens count. \\
    InstEval & lme4 & Free & \citet{lme4}& 73K students 1-5 evaluations of professors from ETH Zurich \\
    Spotify & Tidy Tuesday & Free & \citet{tidytuesday} & 28K songs with their date release, genre, artist, album as well as 12 audio features from which we chose to predict the first one, danceability. \\
    UKB-blood & UK Biobank & Authorized & \citet{UKB} & Subset of 42K UK Biobank with cancer history. To predict triglycerides and other chemicals level in blood we use features such as gender, age, height, weight, skin color and more.\\
    \hline
\end{tabular}
    \caption{Real datasets from \citep{lmmnn_jmlr}: additional details and references.}
    \label{tab:datasets_details}
\end{table}


\end{document}